\documentclass[letterpaper]{article} 
\usepackage{aaai24}  
\usepackage{times}  
\usepackage{helvet}  
\usepackage{courier}  
\usepackage[hyphens]{url}  
\usepackage{graphicx} 
\usepackage{natbib}  
\usepackage{caption} 
\usepackage{algorithm}

\usepackage{appendix}
\usepackage{multirow}
\usepackage{algpseudocode}

\usepackage{makecell}
\usepackage[caption=false,farskip=0pt,labelfont={bf}]{subfig} 
\usepackage{amsmath,amsthm,amssymb}
\usepackage{enumitem}
\usepackage{arydshln}
\usepackage{array}
\usepackage{booktabs}       
\usepackage{url}

\usepackage{amsmath,amsthm,amssymb}
\usepackage{color}

\newtheorem{definition}{Definition}

\newcommand{\beq}{\begin{equation}}
\newcommand{\eeq}{\end{equation}}
\newcommand{\beqs}{\begin{eqnarray}}
\newcommand{\eeqs}{\end{eqnarray}}
\newcommand{\barr}{\begin{array}}
	\newcommand{\earr}{\end{array}}
\newcommand{\bali}{\begin{aligned}}
	\newcommand{\eali}{\end{aligned}}

\newcommand{\Nc}[0]{\ensuremath{\mathcal{N}} }

\newcommand{\Ebb}[0]{\ensuremath{\mathbb{E}} }

\newcommand{\Lbb}[0]{\ensuremath{\mathbb{L}} }

\newcommand{\Rbb}[0]{\ensuremath{\mathbb{R}} }
\newcommand{\Sbb}[0]{\ensuremath{\mathbb{S}} }
\newcommand{\Tbb}[0]{\ensuremath{\mathbb{T}} }

\newcommand{\ie}[0]{\emph{i.e., }}

\newcommand{\eg}[0]{\emph{e.g., }}

\newcommand{\etc}[0]{\emph{etc. }}
\newcommand{\st}[0]{\emph{s.t. }}
\newcommand{\wrt}[0]{\emph{w.r.t. }}

\newcommand{\Amat}[0]{\ensuremath{{\bf A}} }

\newcommand{\Imat}[0]{\ensuremath{{\bf I}} }

\newcommand{\xv}[0]{\ensuremath{\boldsymbol{x}} }
\newcommand{\yv}[0]{\ensuremath{\boldsymbol{y}} }

\newcommand{\Sigmamat}[0]{\ensuremath{\boldsymbol{\Sigma}} }

\newcommand{\Omegamat}[0]{\ensuremath{\boldsymbol{\Omega}}}

\newcommand{\alphav}[0]{\ensuremath{\boldsymbol{\alpha}} }

\newcommand{\thetav}[0]{\ensuremath{\boldsymbol{\theta}} }

\newcommand{\muv}[0]{\ensuremath{\boldsymbol{\mu}} }
\newcommand{\nuv}[0]{\ensuremath{\boldsymbol{\nu}} }

\newcommand{\piv}[0]{\ensuremath{\boldsymbol{\pi}} }

\newcommand{\Dir}[0]{\ensuremath{\mathrm{Dir}} }

\newcommand{\KL}[0]{\ensuremath{\mathrm{KL}} }

\usepackage{newfloat}
\usepackage{listings}
\DeclareCaptionStyle{ruled}{labelfont=normalfont,labelsep=colon,strut=off} 
\floatstyle{ruled}
\newfloat{listing}{tb}{lst}{}
\floatname{listing}{Listing}
 \usepackage{color} 

\title{Big Learning Expectation Maximization}
\author{
	Yulai Cong\thanks{Corresponding author: Yulai Cong.},
	\qquad 
	Sijia Li
}
\affiliations{
	Sun Yat-sen University\\
	yulaicong@gmail.com, lisijia57@163.com

}

\usepackage{bibentry}

\begin{document}

\maketitle

\begin{abstract}
Mixture models serve as one fundamental tool with versatile applications. However, their training techniques, like the popular Expectation Maximization (EM) algorithm, are notoriously sensitive to parameter initialization and often suffer from bad local optima that could be arbitrarily worse than the optimal. To address the long-lasting bad-local-optima challenge, we draw inspiration from the recent ground-breaking foundation models and propose to leverage their underlying big learning principle to upgrade the EM. Specifically, we present the Big Learning EM (BigLearn-EM), an EM upgrade that simultaneously performs joint, marginal, and orthogonally transformed marginal matchings between data and model distributions. Through simulated experiments, we empirically show that the BigLearn-EM is capable of delivering the optimal with high probability; comparisons on benchmark clustering datasets further demonstrate its effectiveness and advantages over existing techniques. The code is available at https://github.com/YulaiCong/Big-Learning-Expectation-Maximization.
\end{abstract}

\section{Introduction}

As a fundamental and prominent tool in statistical machine learning and data science, mixture models are ubiquitously used in versatile practical applications that are associated with density estimation \cite{correia2023continuous}, clustering \cite{chandra2023escaping}, anomaly detection \cite{qu2020anomaly,an2022ensemble}, feature extraction \cite{saire2022global,lin2023real}, model explanation \cite{xie2022joint}, flexible multi-modal prior \cite{saseendran2021shape,lee2021meta}, deblurring \cite{guerrero2007image,yu2011solving}, \etc
Among many variants of mixture models \cite{li2020universal,li2020solving}, the most popular one is the Gaussian Mixture Model (GMM), thanks both to its simplicity and to its capability in approximating any continuous distribution arbitrarily well \cite{lindsay1995mixture,peel2000finite}.
In this paper, we focus on the GMM for presentation, but the presented techniques can be readily extended to other mixture models.

%
%

Although mixture models are widely utilized in practical applications, most of their training techniques are known to be sensitive to parameter initialization \cite{bishop_2006_PRML,jin2016local,kolouri2018sliced}, which alternatively restricts their actual performance. 
For example, the representative Expectation Maximization (EM) algorithm 
has been proven to converge to a bad local optimum that could be arbitrarily worse than the optimal solution with an exponentially high probability, when the number of mixture components exceeds three \cite{jin2016local}.


To address that long-lasting bad-local-optima challenge, we draw inspiration from the recent ground-breaking foundation models, by noticing that they benefit significantly from their massive diverse pretraining tasks, such as mask-and-predict \cite{devlin2018bert,he2022masked} and next-word-prediction \cite{radford2018improving,radford2019language,brown2020language}.
Specifically, \citet{cong2022big} reveal that most of those pretraining strategies actually fall under the big learning principle, \ie leveraging one foundation model to simultaneously and consistently implement many/all joint, conditional, marginal matchings, as well as their transformed matchings, between data and model distributions.

Inspired by that, we propose to leverage the big learning principle to upgrade the conventional EM algorithm to a newly presented Big Learning EM (BigLearn-EM), demonstrating knowledge feedback from cutting-edge foundation models to conventional machine learning.
Specifically, the BigLearn-EM exhaustively exploits its training data with a tailored big learning setup, where joint, marginal, and orthogonally transformed marginal matchings between data and model distributions are simultaneously considered.
On simulated data, the BigLearn-EM delivers the optimal solution with high probability, manifested as an encouraging direction to address the bad-local-optima challenge.



Our contributions are summarized as follows.
\begin{itemize}
    \item We propose the BigLearn-EM, a novel, effective, and easy-to-use algorithm for training mixture models with only EM-type analytical parameter update formulas.
	
    \item We reveal that marginal/conditional matching could help joint matching getting out of bad local optima, which serves as one explanation justifying the successes of foundation models and the big learning principle.
 
    \item Comprehensive clustering experiments are conducted to demonstrate the superiority of the BigLearn-EM and its robustness to the scarcity of training data.
    
	
\end{itemize}


\section{Preliminaries}

We briefly review the preliminaries that lay the foundation of the presented technique, \ie mixture models, the EM algorithm, and the big learning principle.


\subsection{Mixture Models}
\label{sec:mixture_model}

Mixture modeling leverages a mixture (\ie convex combination) of $K$ simple distributions $p_{i}(\xv|\nuv_i)$ with parameters $\nuv_i$ and $i\in\{1,\cdots,K\}$ to construct a more powerful mixture model $p_{\thetav}(\xv)$ for a random variable $\xv\in \Rbb^d$, \ie
\beq
p_{\thetav}(\xv) = \sum\nolimits_{i=1}^K \pi_i p_{i}(\xv|\nuv_i),
\eeq
where the mixture weights $\pi_i>0, \sum\nolimits_{i=1}^K \pi_i = 1$ and $\thetav = \{\pi_i, \nuv_i\}_{i=1}^K$ denotes the model's parameters.

Among various mixture models \cite{li2020solving,li2020universal}, the Gaussian Mixture Model (GMM), also called Mixture of Gaussians (MoG), is the most popular one; its probability density function is
\beq\label{eq:GMM}
p_{\thetav}(\xv) = \sum\nolimits_{i=1}^K \pi_i \Nc(\xv|\muv_i, \Sigmamat_i),
\eeq
where $\muv_i, \Sigmamat_i$ are the mean vector and the covariance
matrix of the $i^{\text{th}}$ Gaussian component, respectively.

\subsection{The Expectation-Maximization Algorithm}
\label{sec:EM}

The Expectation-Maximization (EM) algorithm \cite{dempster1977maximum} is the prominent way of estimating a (Gaussian) mixture model $p_{\thetav}(\xv)$ from a collection of data sampled from an underlying data distribution $q(\xv)$.

Based on the variational inference framework with latent code $z\in\{1,\cdots,K\}$ and an inference arm $q(z|\xv)$ \cite{bishop_2006_PRML,dieng2019reweighted}, the EM algorithm (termed Joint-EM hereafter) maximizes the log-likelihood\footnote{
	In practice, $\Ebb_{q(\xv)}[\cdot]$ is estimated with data samples from $q(\xv)$.
}
\beq\bali\label{eq:MLE}
\Ebb_{q(\xv)}\log p_{\thetav}(\xv) = \Ebb_{q(\xv)} & \Big[
\Ebb_{q(z|\xv)} \log \frac{p_{\thetav}(\xv, z)}{q(z|\xv)}
\\
&  + \KL[q(z|\xv)||p_{\thetav}(z|\xv)] \Big]
\eali\eeq
via alternatively updating $q(z|\xv)$ with an \emph{E-step} and maximizing over $\thetav$ with an \emph{M-step}, that is, 
\beq\bali\label{eq:EM}
\textbf{E-step: }\quad & 
q(z|\xv) = p_{\thetav}(z|\xv)
= \frac{\pi_z \Nc(\xv|\muv_z, \Sigmamat_z)}{\sum\nolimits_{i=1}^K \pi_i \Nc(\xv|\muv_i, \Sigmamat_i)}
\\
\textbf{M-step: }\quad & 
\muv_z = \frac{\Ebb_{q(\xv)} [q(z|\xv) \xv]}{\Ebb_{q(\xv)} [q(z|\xv)]}
\\
&  \Sigmamat_z = \frac{\Ebb_{q(\xv)} [q(z|\xv) (\xv-\muv_z)(\xv-\muv_z)^T]}{\Ebb_{q(\xv)} [q(z|\xv)]}
\\
& \pi_z = \Ebb_{q(\xv)}[q(z|\xv)].
\eali\eeq
Maximizing the log-likelihood in \eqref{eq:MLE} is equivalent to minimizing the Kullback-Leibler (KL) divergence $\KL[q(\xv)||p_{\thetav}(\xv)]$, leading to the KL-based \emph{joint matching} in the joint $\xv$-space, or informally $p_{\thetav}(\xv) \rightarrow q(\xv)$.

\subsection{Big Learning}
\label{sec:biglearn}

Foundation models \cite{stickland2019bert,brown2020language,he2021masked,ramesh2022hierarchical,bao2023one,ChatGPT,ouyang2022training} have demonstrated ground-breaking successes across diverse domains, thanks mainly to their large-scale pretraining on big data.

Observing that the pretraining strategies of foundation models share the similar underlying principle of comprehensively exploiting data information from diverse perspectives, \citet{cong2022big} condenses those strategies into a unified big learning principle that contains most of them as special cases.
Specifically, the big learning leverages one universal model with parameters $\thetav$ to simultaneously match many/all joint, marginal, and conditional data distributions across potentially diverse domains, as defined below.
\begin{definition}[(Uni-modal) big learning \cite{cong2022big}]
	\label{definition:unsupervised_biglearn}
	Given data samples $\xv\in \Rbb^{L}$ from the underlying data distribution $q(\xv)$, the index set $\Lbb=\{1,\cdots,L\}$, and any two {non-overlapping} subsets $\Sbb \subset \Lbb$ and $\Tbb \subseteq \Lbb, \Tbb \neq \emptyset$, the (uni-modal) big learning leverages a universal model $p_{\thetav}(\xv_{\Tbb}|\xv_{\Sbb}), \forall (\Sbb,\Tbb)$ to model many/all joint, conditional, and marginal data distributions simultaneously, \ie 
	\beq\bali\label{eq:uni_model}
	& p_{\thetav}(\xv_{\Tbb}|\xv_{\Sbb})  \longrightarrow q(\xv_{\Tbb}|\xv_{\Sbb}), \forall (\Sbb,\Tbb) \in \Omegamat,
	\eali\eeq
	where $\Omegamat$ is the set that contains the $(\Sbb,\Tbb)$ pairs of interest.
	Given different settings for $(\Sbb,\Tbb)$, $q(\xv_{\Tbb}|\xv_{\Sbb})$ may represent a joint/marginal/conditional data distribution, whose samples are readily available from the training data.
	The actual objective measuring the distance/divergence (or encouraging the matching) between both sides of \eqref{eq:uni_model} should be selected base on the application of interest.
\end{definition}

Based on Remark 3.5 of \citet{cong2022big}, one may alternatively or additionally do big learning in transformed domains, \eg via $p_{\thetav}(\hat \xv_{\Tbb}|\hat \xv_{\Sbb})\longrightarrow q(\hat \xv_{\Tbb}|\hat \xv_{\Sbb})$ with transformation $\hat\xv=g(\xv)$.

Below we will combine the above big learning principle in Definition \ref{definition:unsupervised_biglearn} and Remark 3.5 of \citet{cong2022big} to upgrade the Joint-EM in \eqref{eq:EM} into its big-learning extension, where the universal model $p_{\thetav}(\xv_{\Tbb}|\xv_{\Sbb})$ has an analytical mixture expression for any $(\Sbb,\Tbb)$ pair.


\section{Big Learning Expectation Maximization}
\label{sec:biglearn_EM}

We first reveal a simple but somewhat counter-intuitive fact that lays the foundation of the proposed Big Learning EM (BigLearn-EM) algorithm.
Then, based on that fact and the flexible big learning principle, we design a tailored big-learning task that consists of diverse matchings between data and model distributions. 
Finally, we summarizes and present the easy-to-use BigLearn-EM with only EM-type analytical parameter update formulas.

\subsection{Marginal/Conditional Matching Gets Joint Matching Out of Bad Local Optima}
\label{sec:nonstuck_fact}

\begin{figure*}
	\centering
	\subfloat[]{
		\includegraphics[height=0.4\columnwidth]{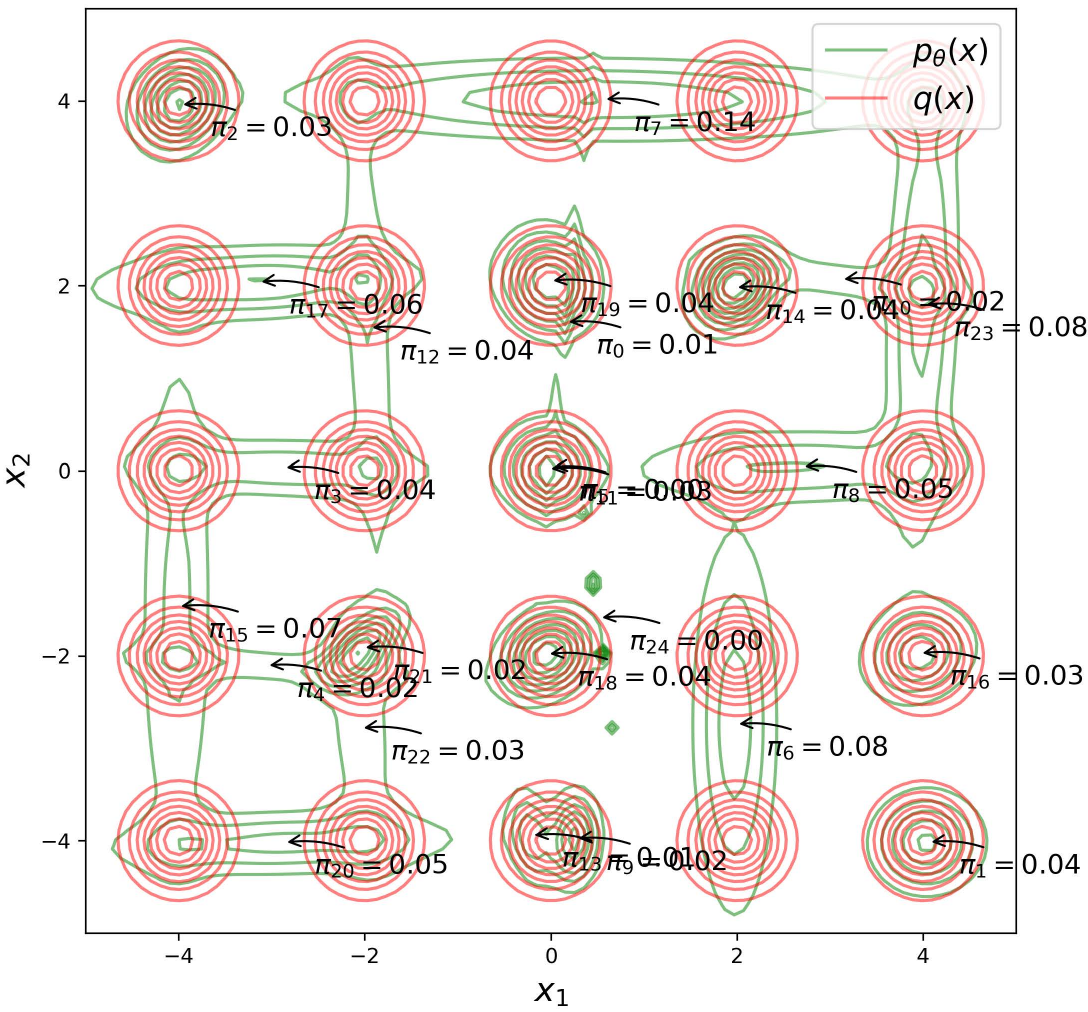}
		\label{fig:jointEMstuck}}
	\subfloat[]{
		\includegraphics[height=0.35\columnwidth]{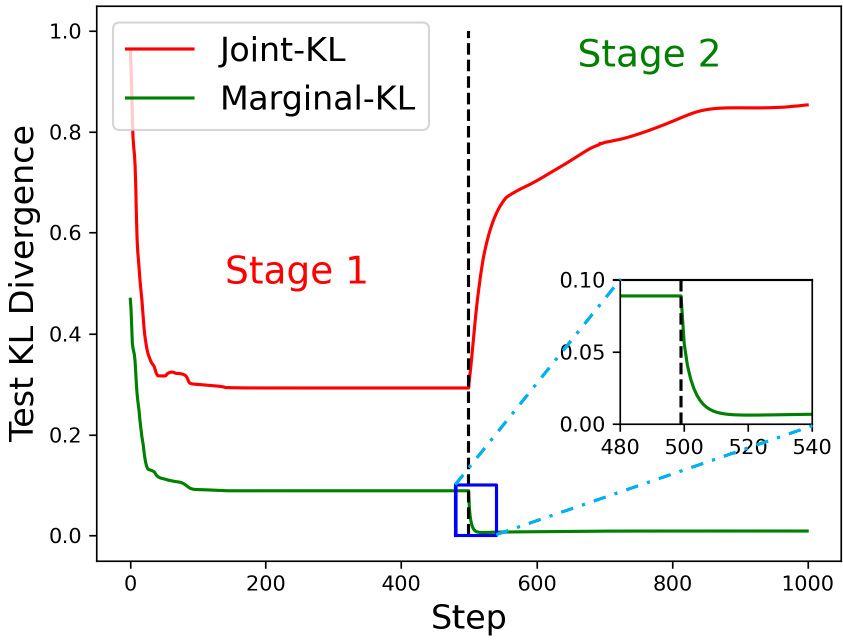}
		\label{fig:MarginEM_notstuck}}
	\subfloat[]{
		\includegraphics[height=0.35\columnwidth]{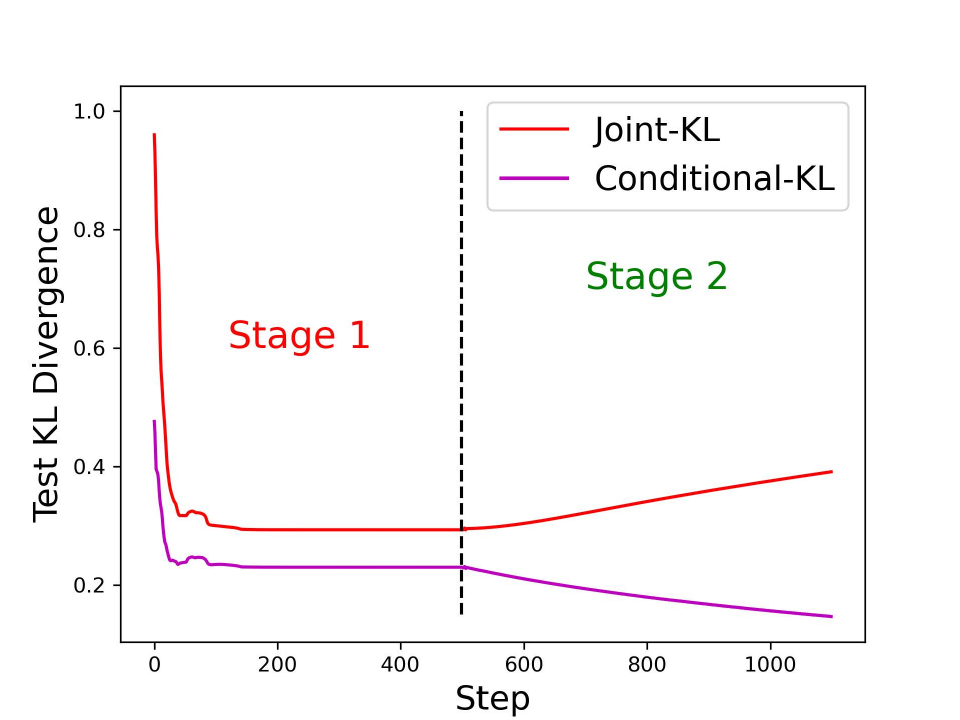}
		\label{fig:Conditional_notstuck}}
	\subfloat[]{
		\includegraphics[height=0.33\columnwidth]{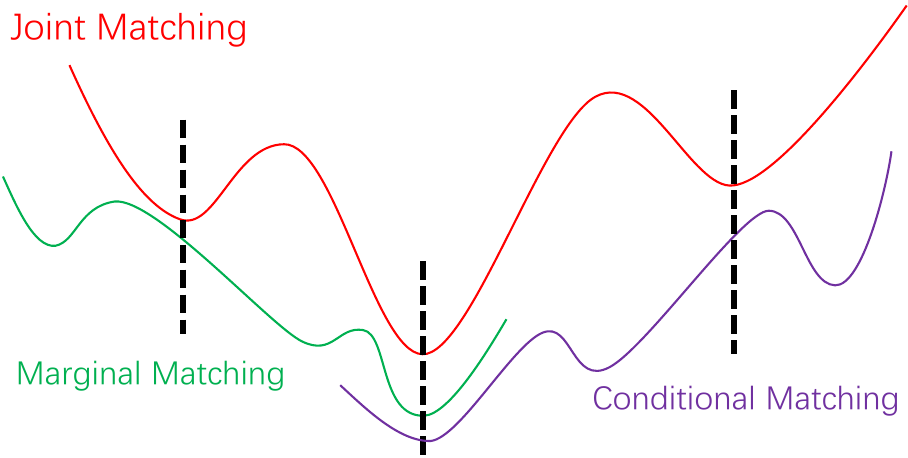}
		\label{fig:digram}}
	\caption{Marginal/Conditional matching gets joint matching out of bad local optima.
		The simulated data distribution $q(\xv)$ is set as a GMM with $25$ components (\ie a $25$-GMM); 
		the model $p_{\thetav}(\xv)$ is also a $25$-GMM with random initialization.
		(a) The Joint-EM of joint matching converges to a bad local optimum. 
		(b) Continuing joint matching in Stage 1, marginal matching in Stage 2 may be further improved and get joint matching out of that bad local optima.
		(c) Similar results are observed when Stage 2 implements conditional matching.
		(d) Schematic diagram of what happens in (b) and (c) from the loss perspective.
	}
	\label{fig:counter_intuitive_fact}
\end{figure*}

It's well-known that the Joint-EM in \eqref{eq:EM} (\ie joint matching $p_{\thetav}(\xv) \rightarrow q(\xv)$) often converges to a bad local optimum that could be arbitrarily worse than the optimal with an exponentially high probability \cite{bishop_2006_PRML,jin2016local,kolouri2018sliced}, when the number of mixture components exceeds three.
Fig. \ref{fig:jointEMstuck} illustrates an example bad local optimum when implementing the Joint-EM on simulated data sampled from a GMM with $25$ components (abbreviated as $25$-GMM hereafter).

Next, with notations $\xv\in \Rbb^{L}$ and the index set $\Lbb=\{1,\cdots,L\}$, let's consider the relationships among joint matching with $p_{\thetav}(\xv) \rightarrow q(\xv)$, marginal matching with $p_{\thetav}(\xv_{\Tbb}) \rightarrow q(\xv_{\Tbb})$, and conditional matching with $p_{\thetav}(\xv_{\Tbb}|\xv_{\Sbb}) \rightarrow q(\xv_{\Tbb}|\xv_{\Sbb})$, 
where $\Tbb \subseteq \Lbb$, $\Tbb \neq \emptyset$, $\Sbb \subset \Lbb$, $\Sbb \cap \Tbb = \emptyset$, and $\xv_{\Tbb}$ is the marginal vector of $\xv$ indexed by $\Tbb$.

Intuitively, one may anticipate that performing joint matching (with \eg the Joint-EM in \eqref{eq:EM}) will automatically lead to the convergences of both marginal matching (with \eg the following Marginal-EM in \eqref{eq:marginal_EM}) and conditional matching (via \eg maximizing the follow-up conditional log-likelihood in \eqref{eq:conditional_MLE}). 
\beq\resizebox{\linewidth}{!}{$
\bali\label{eq:marginal_EM}
& \textbf{Marginal Matching} \quad \Ebb_{q(\xv_{\Tbb})}\log p_{\thetav}(\xv_{\Tbb}) 
\\
& \,\,\,\textbf{E-step: } 
q(z|\xv_{\Tbb}) = p_{\thetav}(z|\xv_{\Tbb})
= \frac{\pi_z \Nc(\xv_{\Tbb}|\muv_{z\Tbb}, \Sigmamat_{z\Tbb\Tbb})}{\sum\nolimits_{i=1}^K \pi_i \Nc(\xv_{\Tbb}|\muv_{i\Tbb}, \Sigmamat_{i\Tbb\Tbb})}
\\
& \,\,\,\textbf{M-step: }
\muv_{z\Tbb} = \frac{\Ebb_{q(\xv_{\Tbb})} [q(z|\xv_{\Tbb}) \xv_{\Tbb}]}{\Ebb_{q(\xv_{\Tbb})} [q(z|\xv_{\Tbb})]}
\\
& \qquad\qquad \Sigmamat_{z\Tbb\Tbb} = \frac{\Ebb_{q(\xv_{\Tbb})} [q(z|\xv_{\Tbb}) (\xv_{\Tbb}-\muv_{z\Tbb})(\xv_{\Tbb}-\muv_{z\Tbb})^T]}{\Ebb_{q(\xv_{\Tbb})} [q(z|\xv_{\Tbb})]}
\\
& \qquad\qquad \pi_z = \Ebb_{q(\xv_{\Tbb})}[q(z|\xv_{\Tbb})],
\eali
$}
\eeq
where $\muv_{z\Tbb}$ and $\Sigmamat_{z\Tbb\Tbb}$ represent the $\Tbb$-indexed marginal vector/matrix of $\muv_{z}$ and $\Sigmamat_{z}$, respectively.
\beq\bali\label{eq:conditional_MLE}
\textbf{Conditional Matching} \quad
\Ebb_{q(\xv_{\Sbb})q(\xv_{\Tbb} | \xv_{\Sbb})} {\log} p_{\thetav}(\xv_{\Tbb} | \xv_{\Sbb})
\eali\eeq

However, we empirically reveal below that the above intuition will \emph{not} hold true when joint matching (or Joint-EM) gets stuck at a bad local optimum.

Specifically, we conduct two-stage experiments on simulated data from a $25$-GMM $q(\xv)$ (see Fig. \ref{fig:jointEMstuck}), where the model $p_{\thetav}(\xv)$ is also a $25$-GMM with random initialization\footnote{
	Different from prior methods initializing parameters $\{\muv_i\}_{i=1}^K$ with uniformly sampled training data, we use the more challenging Gaussian random initialization for $\{\muv_i\}_{i=1}^K$ to highlight the power of the proposed BigLearn-EM.
},
Stage 1 implements joint matching (with Joint-EM in \eqref{eq:EM}), and, directly following Stage 1, Stage 2 either implements marginal matching (with Marginal-EM in \eqref{eq:marginal_EM}) or conditional matching (via maximizing the conditional log-likelihood in \eqref{eq:conditional_MLE} with gradient accent).

Fig. \ref{fig:counter_intuitive_fact} demonstrates the results.
It's clear from Fig. \ref{fig:jointEMstuck} that joint matching gets stuck at a bad local optimum.
As shown in Fig. \ref{fig:MarginEM_notstuck}, the convergence of joint matching in Stage 1 does not necessarily result in the convergence of marginal matching, because continually performing Marginal-EM in Stage 2 further improves marginal matching.
Similar phenomena are observed in Fig. \ref{fig:Conditional_notstuck} for conditional matching.
That means bad local optima where joint matching gets stuck are not local optima for marginal/conditional matching, as illustrated in the left and right dashed lines of the schematic diagram in Fig. \ref{fig:digram}.
Alternatively, that inconsistency among joint, marginal, and conditional matchings may be leveraged, \eg to detect bad local optima of each matching or to help each other get out of bad local optima.

It's worth highlighting that the center dashed line in Fig. \ref{fig:digram} is located at a \emph{consistent local optimum} for joint, marginal, and conditional matchings; more importantly, that consistency property is what the optimal solution must satisfy. 
The above analysis serves as an example justification for simultaneous joint, marginal, and conditional matchings, \ie the big learning principle in \eqref{eq:uni_model} that underlies most successful foundation models.

\subsection{On Tailoring a Big-Learning Task to Produce an Easy-To-Use  BigLearn-EM}
\label{sec:orth_margin_biglearnEM}

Based on what's revealed in the previous section, one may naively follow the vanilla big learning principle in \eqref{eq:uni_model} to conduct multitasking joint, marginal, conditional matchings in the  original $\xv$-space, \ie 
\beq\bali\label{eq:biglearn_x}
\max\nolimits_{\thetav} \Ebb_{q(\Sbb, \Tbb)} \Ebb_{q(\xv_{\Sbb})q(\xv_{\Tbb} | \xv_{\Sbb})} {\log} p_{\thetav}(\xv_{\Tbb} | \xv_{\Sbb}),
\eali\eeq
where $q(\Sbb, \Tbb)$ represents the sampling process of $(\Sbb, \Tbb)$. Note $q(\Sbb, \Tbb)$ actually determines the relative weightings among joint, marginal, and conditional matchings.
However, 
it's not easy to design EM-type analytical update formulas for conditional matching in \eqref{eq:conditional_MLE}, even though such formulas are readily available for both joint and marginal matchings, as given in \eqref{eq:EM} and \eqref{eq:marginal_EM}, respectively.

To avoid a hybrid algorithm that contain both EM-type and gradient accent updates and thus may not easy to use, we leverage the flexibility of big learning discussed in Remark 3.5 of \citet{cong2022big} to further combine marginal matchings in randomly transformed $\yv$ domains with the joint and marginal matchings in the original $\xv$ domain, to form the tailored big-learning task.

Specifically, we employ orthogonal transformations $\yv=\Amat\xv$, where $\Amat$ is a randomly sampled orthogonal matrix. 
Correspondingly, the transformed training data $\yv\sim \bar q_{\Amat}(\yv)$ are generated via $\yv=\Amat\xv, \xv\sim q(\xv)$, the model in a transformed domain $\bar p_{\thetav, \Amat}(\yv)$ is also a GMM with the analytical expression of 
\beq\bali\label{eq:tran_model}
\bar p_{\thetav, \Amat}(\yv) & = p_{\thetav}(\xv) \big| \frac{\partial \xv}{\partial \yv} \big| 
= \sum\nolimits_{i=1}^K \pi_i \Nc(\yv|\bar\muv_i, \bar\Sigmamat_i),
\eali\eeq
where $\bar\muv_i=\Amat\muv_i$, $\bar\Sigmamat_i=\Amat\Sigmamat_i\Amat^T$, and the transformed marginal matching has EM-type analytical update formulas 
\beq
\resizebox{\linewidth}{!}{$
\bali\label{eq:marginal_EM_y}
& \textbf{\makecell{Randomly Transformed\\Marginal Matching}} \quad 
\Ebb_{\bar q_{\Amat}(\yv_{\Tbb})} \log \bar p_{\thetav, \Amat}(\yv_{\Tbb}) 
\\
& \bali
\textbf{E-step: } & 
\bar q_{\Amat}(z|\yv_{\Tbb}) = \bar p_{\thetav, \Amat}(z|\yv_{\Tbb})
= \frac{\pi_z \Nc(\yv_{\Tbb}|\bar \muv_{z\Tbb}, \bar \Sigmamat_{z\Tbb\Tbb})}{\sum\nolimits_{i=1}^K \pi_i \Nc(\yv_{\Tbb}|\bar \muv_{i\Tbb}, \bar \Sigmamat_{i\Tbb\Tbb})}
\\
\textbf{M-step: } & 
\bar \muv_{z\Tbb} = \frac{\Ebb_{\bar q_{\Amat}(\yv_{\Tbb})} [\bar q_{\Amat}(z|\yv_{\Tbb}) \yv_{\Tbb}]}{\Ebb_{\bar q_{\Amat}(\yv_{\Tbb})} [\bar q_{\Amat}(z|\yv_{\Tbb})]}
\\
& \bar \Sigmamat_{z\Tbb\Tbb} = \frac{\Ebb_{\bar q_{\Amat}(\yv_{\Tbb})} [\bar q_{\Amat}(z|\yv_{\Tbb}) (\yv_{\Tbb}-\bar\muv_{z\Tbb})(\yv_{\Tbb}-\bar\muv_{z\Tbb})^T]}{\Ebb_{\bar q_{\Amat}(\yv_{\Tbb})} [\bar q_{\Amat}(z|\yv_{\Tbb})]}
\\
& \pi_z = \Ebb_{\bar q_{\Amat}(\yv_{\Tbb})}[\bar q_{\Amat}(z|\yv_{\Tbb})]
\\
\textbf{Update } & \thetav\textbf{: } 
\muv_{z} = \Amat^T \bar \muv_{z}^{'},
\quad 
\Sigmamat_{z} = \Amat^T \bar \Sigmamat_{z}^{'} \Amat,
\eali
\eali
$}\eeq
where $\bar\muv_{z}^{'} / \bar \Sigmamat_{z}^{'}$ is the $\Tbb$-partially updated $\bar\muv_{z} / \bar \Sigmamat_{z}$ after the M-step.
Note any joint matching in the transformed $\yv$ domain will deliver the same update formulas as in \eqref{eq:EM}.

To summarize, the tailored big-learning task contains three kinds of matching, that is, joint, marginal, and transformed marginal matchings, 
each of which has EM-type analytical formulas for parameter updates, \ie \eqref{eq:EM}, \eqref{eq:marginal_EM}, and \eqref{eq:marginal_EM_y}, respectively.

\subsection{Finalizing the BigLearn-EM}
\label{sec:final_biglearnEM}

Before finalizing our BigLearn-EM, an issue of the EM-type updates should be addressed.
It's easy to verify that, during the EM iterations, once a mixture weight $\pi_z$ becomes zero, it stays zero thereafter. 
Empirically, this issue hinders the EM-type updates in \eqref{eq:EM}, \eqref{eq:marginal_EM}, and \eqref{eq:marginal_EM_y} from making full use of the available mixture components, even though the occupied components have no enough modeling capacity.

To address that issue, we leverage the Maximum a posteriori (MAP) estimate in place of the vanilla maximum log-likelihood estimate on the mixture weights $\piv$ following \citet{bishop_2006_PRML}. 
Accordingly, taking Joint-EM in \eqref{eq:EM} as an example, the update rule for $\piv$ is replaced by
\beq\label{eq:post_pi}
\pi_z = \frac{\Ebb_{q(\xv)} [q(z|\xv)] + \eta}{1 + K \eta},	
\eeq
where $\eta >0$ is a small constant.
Similar modifications are also applied to \eqref{eq:marginal_EM} and \eqref{eq:marginal_EM_y}, respectively. 
Detailed derivations are given in Appendix \ref{appsec:MAP_EM}.

Based on the aforementioned tailored big-learning task and the MAP modification on $\piv$, we finalize the training objective of the BigLearn-EM as 
\beq\bali\label{eq:biglearn_y}
\max\nolimits_{\thetav} \,\, & \Ebb_{q(\Sbb, \Tbb)q(\Amat)} \Ebb_{\bar q_{\Amat}(\yv_{\Sbb})\bar q_{\Amat}(\yv_{\Tbb} | \yv_{\Sbb})} {\log} \bar p_{\thetav,\Amat}(\yv_{\Tbb} | \yv_{\Sbb})
\\
& + \gamma\log p_{\alpha}(\piv),
\eali\eeq
where $q(\Sbb, \Tbb)$ and $q(\Amat)$ represent the sampling process of $(\Sbb, \Tbb)$ and the orthogonal matrix $\Amat$, respectively.
$p_{\alpha}(\piv)$ is the prior for $\piv$. $\gamma$ is a hyper-parameter. 
Joint/Marginal matching may be recovered with $\Sbb=\emptyset,\Amat=\Imat$.

Algorithm \ref{alg:biglearn_EM} summarizes the presented BigLearn-EM, where only easy-to-use EM-type updates are employed.
We associate the stopping criterion with ``mixing'' following the MCMC literature because of the random implementation of joint, marginal, or transformed marginal matchings; in the experiments, we run Algorithm \ref{alg:biglearn_EM} for a fixed number of iterations.
Besides, it's worth highlighting that the BigLearn-EM can naturally handle incomplete data (via its marginal matchings) thanks to its big learning nature.

\begin{algorithm}[tb]
	\caption{Big Learning Expectation Maximization} 
	\label{alg:biglearn_EM}
	\begin{algorithmic}[1]
		\Require Training data, the number $K$ of mixture components, probabilities $[P_1,P_2]$ for joint and marginal matchings, and the number $W$ of local updates.
		\Ensure A consistent local optimum $\thetav^{*} = \{\pi_i^{*}, \muv_i^{*}, \Sigmamat_i^{*}\}_{i=1}^K$.  
		\State Randomly initialize $\thetav = \{\pi_i, \muv_i, \Sigmamat_i\}_{i=1}^K$
		\While{Not Mixing}
		\State With probability $P_1$, 
		\State \qquad do Joint-EM with \eqref{eq:EM}/\eqref{eq:post_pi} for $W$ iterations 
		\State With probability $P_2$, 
		\State \qquad ($i$) uniformly sample an index subset $\Tbb$, and
		\State \qquad ($ii$) do Marginal-EM with \eqref{eq:marginal_EM}/\eqref{eq:post_pi} for $W$ iters 
		\State With probability $1-P_1-P_2$, 		
		\State \qquad ($i$) uniformly sample an orthogonal matrix $\Amat$
		\\ \Comment{\texttt{scipy.stats.ortho\_group}}	
		\State \qquad ($ii$) uniformly sample an index subset $\Tbb$, and
		\State \qquad ($iii$) do Transformed Marginal-EM with
		\\ \qquad\qquad\quad\, \eqref{eq:marginal_EM_y}/\eqref{eq:post_pi} for $W$ iterations
		\EndWhile
	\end{algorithmic}
\end{algorithm}

\section{Related Work}

\textbf{Analysis and Improvements of the EM Algorithm} 
In general settings, the (Joint-)EM algorithm only have local convergence guarantee, that is, it converges to the optimal only if the parameters are initialized within a close neighborhood of that optimal \cite{yan2017convergence,zhao2020statistical,balakrishnan2017statistical}.
Although \citet{xu2016global,daskalakis2017ten,qian2019global} have established the global convergence for Joint-EM on learning GMMs with two components, a global convergence guarantee is generally impossible for GMMs with $K\ge3$ components, where Joint-EM converges to a bad local optimum with an exponentially high probability \cite{jin2016local}. 
To deal with that bad-local-optima challenge, many efforts have been made to improve Joint-EM, most of which focus on clever parameter initialization, seeking to help Joint-EM bypass bad local optima before E-M iterates \cite{bachem2016approximate,bachem2016fast,scrucca2016mclust,bachem2018scalable,exarchakis2022sampling,tobin2023reinforced}.
By contrast, the proposed BigLearn-EM, with random initialization, directly tackle the bad-local-optima challenge with diverse joint, marginal, transformed marginal EM updates, empirically delivering boosted performance than the Joint-EM (see the experiments).



\textbf{Other Methods for Learning Mixture Models}
Besides the popular EM algorithm, many other methods for learning GMMs have also been developed based on, \eg Markov chain Monte Carlo (MCMC) \cite{rasmussen1999infinite,favaro2013MCMC,das2014collapsed}, moments \cite{ge2015learning,kane2021robust,pereira2022tensor}, adversarial learning \cite{lin2018pacgan,farnia2023gat}, and optimal transport \cite{kolouri2018sliced,li2020solving,yan2023learning}.
Specifically, the SW-GMM \cite{kolouri2018sliced} leverages the Radon transform to randomly project the high-dimensional GMM learning task into \emph{one-dimensional} sliced subspace, where the sliced Wasserstein distance between the projected data and model distributions is minimized. However, the computational complexity of the SW-GMM grows exponentially as the number of dimensions, rendering it unsuitable for modeling high-dimensional data \cite{li2020solving,deshpande2019max,kolouri2019generalized}.
Different from the aforementioned methods resorting to expensive moment-matching, unstable adversarial learning, or complicated Wasserstein distances, the presented BigLearn-EM is both easy-to-understand and easy-to-use, since it's a direct big-learning upgrade of the EM algorithm with only EM-type analytical formulas for parameter updates (and thus the same computational complexity as that of the EM).




%
%
%

\section{Experiments}
\label{sec:Exp}

We first present the detailed ablation study that produces the BigLearn-EM from the vanilla Joint-EM. 
Then, we demonstrate the effectiveness of the BigLearn-EM in comprehensive real-world clustering applications.
Finally, modified clustering experiments are conducted to reveal its robustness to data scarcity. 


\subsection{Ablation Study That Produces the BigLearn-EM}
\label{sec:Ablation}

Based on the $25$-GMM simulation setup in Fig. \ref{fig:counter_intuitive_fact}, we first present the detailed ablation study that produces the BigLearn-EM in Algorithm \ref{alg:biglearn_EM}.
Specifically, we start with the Joint-EM in \eqref{eq:EM} and test the performance when gradually introducing additional MAP estimate for $\piv$ (marked as ``+Pr''), Marginal Matching in \eqref{eq:marginal_EM} (``+MM''), Conditional Matching in \eqref{eq:conditional_MLE} (``+CM''), and Randomly Transformed Marginal Matching in \eqref{eq:marginal_EM_y} (``+RTMM'') with different number $W$ of local updates in Algorithm \ref{alg:biglearn_EM} (marked as ``+W'').

The results from $10$ different runs (with different random seeds) are summarized in Table \ref{tab:ablation_study}, where introducing prior for $\piv$ (\ie ``+Pr'') improves the test joint KL divergence by $14.4\%$ on average, despite with a doubly worsened standard deviation. By additionally employing marginal/conditional matching (\ie ``+MM$/$+CM''), both the mean and standard deviation improve steadily, highlighting the benefits of the implicit diverse inter-regularization among various learning objectives of big learning.
Further, boosted performance emerges from employing the Randomly Transformed Marginal Matching (\ie ``+RTMM''), thanks to its significantly expanded diversity of matching, highlighting the effectiveness of the big learning principle as well as the importance of the diversity of big-learning tasks.

\begin{table}[tb]
	\centering
	\caption{Ablation study on the $25$-GMM simulated datasets. 
		``+Pr'' means employing the MAP estimate for $\piv$ with \eqref{eq:post_pi}.
		``+MM$/$+CM$/$+RTMM'' means introducing additional Marginal Matching, Conditional Matching, and Randomly Transformed Marginal Matching, respectively.
		``+W5'' indicates employing $W=5$ local updates in Algorithm \ref{alg:biglearn_EM}.
	}
	\resizebox{\columnwidth}{!}{
		\begin{tabular}{l c c}
			\hline\hline 
			\qquad\qquad\multirow{3}{*}{Method}  &   \multicolumn{2}{c}{Test Joint KL Divergence}
			\\ \cline{2-3}
			& Mean & \makecell{Standard\\Deviation}
			\\ \hline\hline
			Joint-EM &  $0.263$ & $0.035$
			\\ \hline \hline
			\, + Pr & $0.225$ & $0.073$
			\\
			\, + Pr + MM & $0.141$ & $0.054$ 
			\\
			\, + Pr + MM + CM & $0.124$ & $0.044$
			\\ \hline
			\, + Pr + MM + RTMM + W$1$ & $0.077$ & $0.034$ 
			\\
			\makecell{+ Pr + MM + RTMM + W$5$\\{\it (BigLearn-EM)}} & $\bf 0.030$ & $\bf 0.006$ 
			\\
			\, + Pr + MM + RTMM + W$10$ & $0.031$ & $0.007$ 
			\\ \hline \hline
		\end{tabular}
		}
	\label{tab:ablation_study}
\end{table}

\begin{figure*}[htbp]
	\centering
	\subfloat[]{
		\includegraphics[width=1.4\columnwidth]{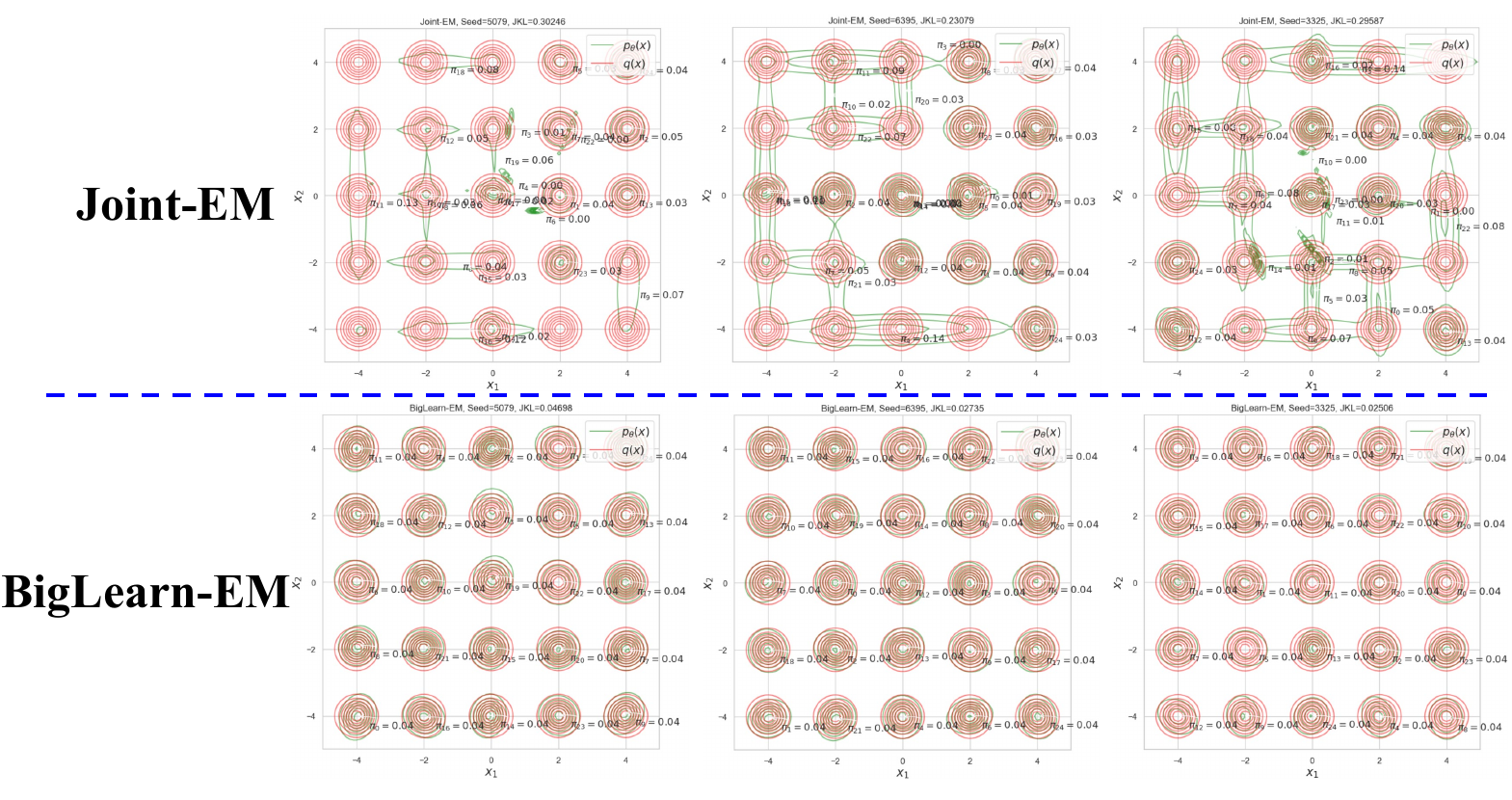}
		\label{fig:JointEM_BiglearnEM_explicit}}
    \begin{minipage}[b]{0.65\columnwidth}
		\centering
		\subfloat[]{
		\includegraphics[width=0.9\columnwidth]{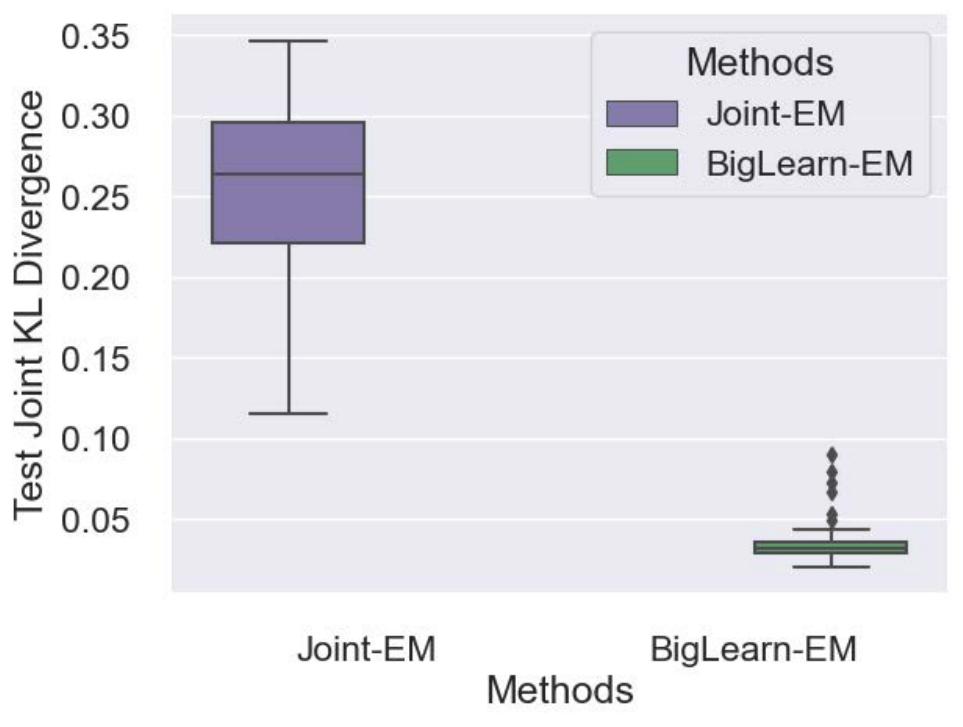}
		\label{fig:JointEM_BiglearnEM_boxplot}}
        \\
        \subfloat[]{
        \includegraphics[width=0.7\columnwidth]{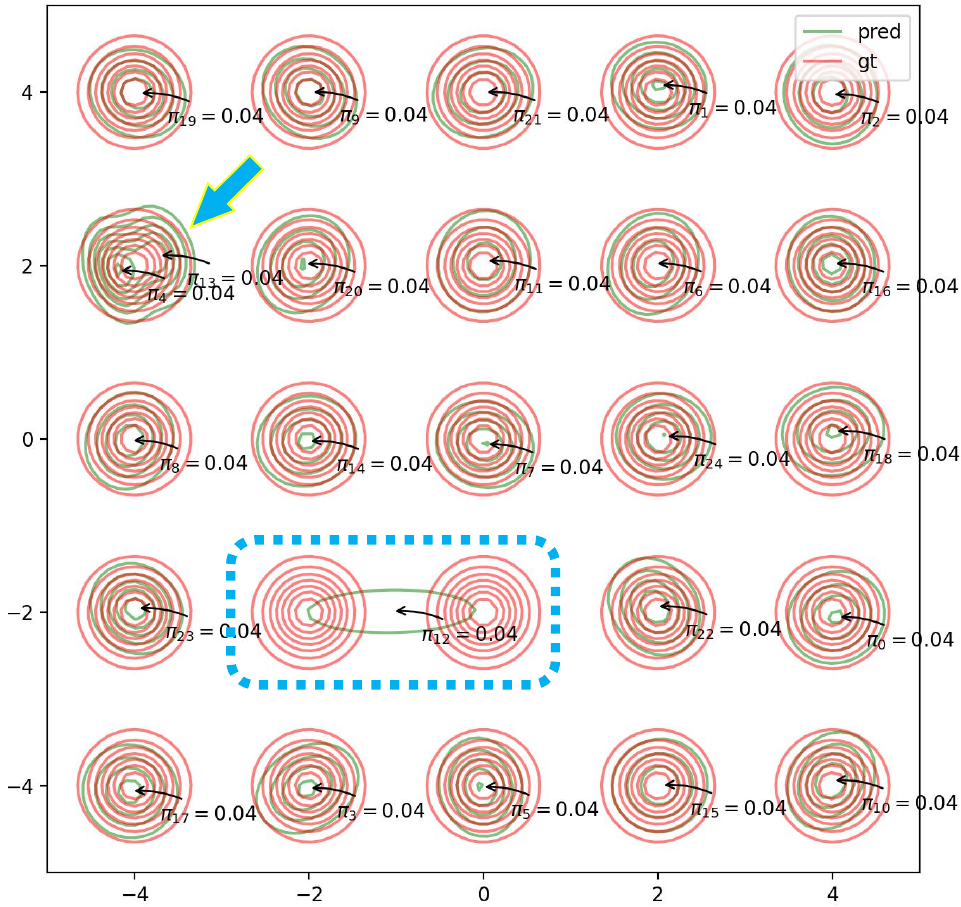}
		\label{fig:to_improve}}
	\end{minipage}
	\caption{Comparisons between the Joint-EM and the BigLearn-EM.	
		(a) Explicit demonstrations of the local optima from both methods \wrt different random seeds of $5079$, $6395$, and $3325$, respectively.
		(b) Boxplot of the test joint KL divergences from $100$ runs of both methods with different random seeds.
		(c) An example state where the BigLearn-EM wanders around for many iterations.
	}
	\label{fig:JointEM_BiglearnEM}
\end{figure*}

For explicit comparisons between the Joint-EM and the BigLearn-EM, Fig. \ref{fig:JointEM_BiglearnEM_explicit} demonstrates the local optima where both methods converge. 
As expected, Joint-EM fails to make full use of the available $25$ mixture components, suffering from bad local optima that could be arbitrarily worse than the optimal solution \cite{jin2016local}. 
By contrast, the presented BigLearn-EM, thanks to its big-learning nature, manages to fully exploit the $25$ mixture components by placing each component to one data mode, delivering global optima with high probability in this simulation (refer to Fig. \ref{fig:JointEM_BiglearnEM_boxplot}).
By considering that the BigLearn-EM merely uses the Gaussian random initialization for $\{\muv_i\}_{i=1}^K$, it's therefore interesting to theoretically verify whether big learning could contribute to a global convergence guarantee for GMMs with $K\ge3$ components; we leave that as future research.



\begin{table*}[p]
\centering
\caption{Clustering performance on real-world benchmark datasets.
    All the compared methods share the same settings for the GMM model $p_{\thetav}(\xv)$. 
    The results are calculated based on $100$ runs with different random seeds.
    Higher is better for all metrics.
    }
 \resizebox{\textwidth}{!}{
    \begin{tabular}{ccccccc}
			\hline \hline
			Dataset                    &       Metric         & K-Means      	    & WM-GMM 		   & SW-GMM			        & Joint-EM 		          & BigLearn-EM           \\ \hline 
			\multirow{3}{*}{Connect-4} & NMI            & $0.0025\pm0.0014$& $0.0016\pm0.0084$& $0.0016\pm0.0097$& $\bf0.0028\pm0.0015$& $0.0022\pm0.0012$\\
			& ARI            & $0.0003\pm0.0011$& $0.0003\pm0.0042$& $0.0005\pm0.0047$& $0.0013\pm0.0049$& $\bf0.0017\pm0.0031$\\
			& Joint-LL & \textemdash & $88.568\pm3.45$& $85.659\pm7.498$& $91.254\pm4.8061$& $\bf95.324\pm3.2022$\\ \hline
			\multirow{3}{*}{Covtype}   & NMI            & $0.145\pm0.0647$& $0.101\pm0.0158$& $0.138\pm0.0341$& $0.119\pm0.0272$& $\bf0.171\pm0.0143$\\
			& ARI            & $0.032\pm0.0147$& $0.065\pm0.0497$& $0.037\pm0.0658$& $0.057\pm0.0216$& $\bf0.070\pm0.0151$\\
			& Joint-LL & \textemdash & $70.957\pm0.059$& $70.268\pm1.632$& $72.194\pm1.9089$& $\bf74.002\pm0.8828$\\ \hline
			\multirow{3}{*}{Glass}     & NMI            & $0.431\pm0.0567$& $0.419\pm0.0726$& $0.426\pm0.0534$& $0.436\pm0.0644$& {$\bf0.459\pm0.0440$}    \\
			& ARI            & $0.164\pm0.0758$& $0.198\pm0.0511$& $0.195\pm0.0479$& {$0.220\pm0.0526$}& $\bf0.228\pm0.0423$\\
			& Joint-LL & \textemdash & $7.142\pm0.8023$& {$\bf7.148\pm0.9546$}& $7.008\pm1.0364$& $7.140\pm1.0025$\\ \hline
			\multirow{3}{*}{Letter}    & NMI            & $0.368\pm0.0064$& $0.279\pm0.0033$& $0.478\pm0.0186$& $0.492\pm0.0169$& $\bf0.532\pm0.0121$\\
			& ARI            & $0.130\pm0.0056$& $0.012\pm0.0032$& $0.190\pm0.0202$& $0.193\pm0.0181$& $\bf0.244\pm0.0165$\\
			& Joint-LL & \textemdash & $12.38\pm0.1024$& $19.045\pm0.1548$& $19.297\pm0.1877$& $\bf19.664\pm0.1404$\\ \hline
			\multirow{3}{*}{Pendigits} & NMI            & $0.716\pm0.0059$& $0.782\pm0.0233$& $0.744\pm0.0385$& $0.771\pm0.0323$& $\bf0.823\pm0.0202$\\
			& ARI            & $0.596\pm0.0180$& $0.679\pm0.0487$& $0.600\pm0.0663$& $0.626\pm0.0622$& $\bf0.724\pm0.0392$\\
			& Joint-LL & \textemdash & $10.068\pm0.1824$& $9.870\pm0.2198$& $9.960\pm0.2545$& $\bf10.266\pm0.0979$\\ \hline
			\multirow{3}{*}{Satimage}  & NMI            & $0.586\pm0.0012$& $0.575\pm0.0256$& $0.598\pm0.0542$& $0.587\pm0.0311$& $\bf0.617\pm0.0291$\\
			& ARI            & $0.487\pm0.0007$& $0.498\pm0.0451$& $0.505\pm0.1562$& $0.470\pm0.0765$& $\bf0.527\pm0.0621$\\
			& Joint-LL & \textemdash & $39.214\pm0.0035$& $39.384\pm0.092$& $39.387\pm0.0062$& $\bf39.430\pm0.1109$\\ \hline
			\multirow{3}{*}{Seismic}   & NMI            & $0.121\pm0.0015$& $0.167\pm0.0145$& $0.196\pm0.0090$& $0.198\pm0.0259$& $\bf0.212\pm0.0243$\\
			& ARI            & $0.106\pm0.0033$& $0.113\pm0.1584$& $0.0892\pm0.0426$& $0.057\pm0.0292$& $\bf0.129\pm0.0265$\\
			& Joint-LL & \textemdash & $41.958\pm0.2185$& $42.234\pm0.1441$& $42.050\pm0.8780$& $\bf42.449\pm0.8896$\\ \hline
			\multirow{3}{*}{Svmguide2} & NMI            & $0.105\pm0.0504$& $0.098\pm0.0372$& $0.108\pm0.0638$& $0.085\pm0.0746$& $\bf0.196\pm0.0722$\\
			& ARI            & $0.087\pm0.0576$& $0.061\pm0.0348$& $0.087\pm0.0911$& $0.050\pm0.0820$& $\bf0.206\pm0.0869$\\
			& Joint-LL & \textemdash & $10.248\pm0.0546$& $\bf10.416\pm0.4158$& $10.404\pm0.4240$& $10.371\pm0.3993$\\ \hline
			\multirow{3}{*}{Vehicle}   & NMI            & $0.169\pm0.0282$& $0.218\pm0.0152$& $0.178\pm0.0545$& $0.197\pm0.0655$& $\bf0.249\pm0.0641$\\
			& ARI            & $0.089\pm0.0250$& $0.102\pm0.0131$& $0.085\pm0.0533$& $0.094\pm0.0476$& $\bf0.131\pm0.0471$\\
			& Joint-LL & \textemdash & $22.2998\pm1.0494$& $22.473\pm1.0635$& $22.896\pm1.3036$& $\bf23.738\pm1.1594$\\ \hline\hline
		\end{tabular}
 }
\label{table:clustering}
\end{table*}

\subsection{BigLearn-EM for Real-World Data Clustering}

Clustering stands as a representative application of GMM, addressing the task of categorizing unlabeled data into coherent and distinct clusters.

To validate the effectiveness of the BigLearn-EM in real-world clustering applications, we conduct comprehensive experiments on diverse clustering datasets, including Connect-4, Covtype, Glass, Letter, Pendigits, Satimage, Seismic, Svmguide2, and Vehicle (see Appendix \ref{appsec:Exp_settings_cluster} for details). 
The BigLearn-EM is systematically benchmarked against representative established clustering techniques, \ie 
the K-Means \cite{bottou1994convergence}, the SW-GMM \cite{kolouri2018sliced}, and the WM-GMM \cite{li2020solving}, and the Joint-EM algorithm. 
Three testing metrics are adopted for performance evaluation, including 
\begin{enumerate}
    \item the normalized mutual information (NMI) \cite{strehl2002cluster}, which quantifies how much the predicted clustering is informative about the true labels;
    \item the adjusted rand index (ARI) \cite{hubert1985comparing,steinley2004properties}, which measures the agreement between an estimated clustering and a reference clustering; and
    \item the test joint log-likelihood (Joint-LL), which reflects how well the learned model describes the testing data from the joint KL divergence perspective.
\end{enumerate}


\begin{figure}[tbp]
	\centering
	\includegraphics[width=0.95\columnwidth]{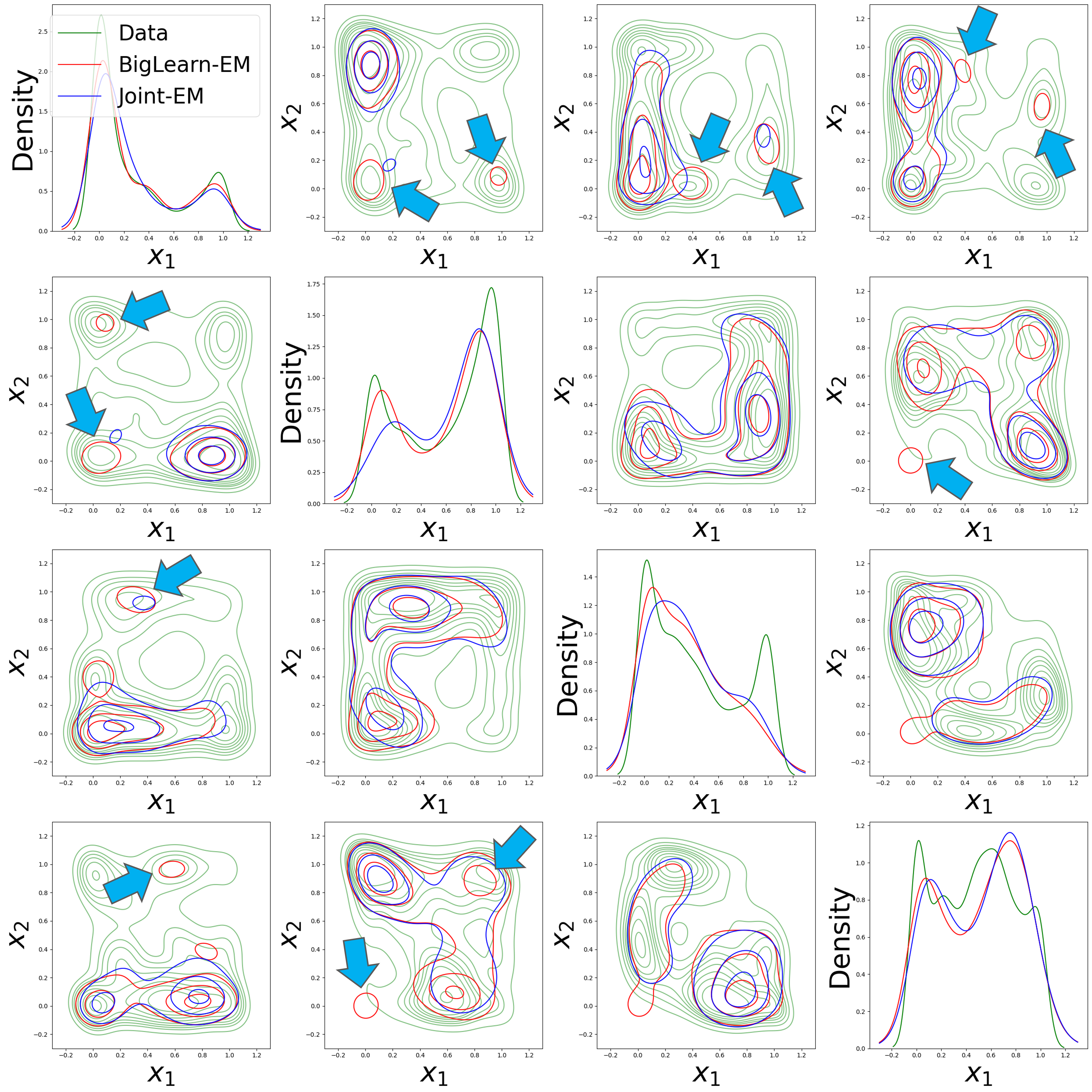}
	\caption{Demonstration of the transformed marginal matchings associated with the top four principal components on the Pendigits dataset.	
	}
	\label{fig:marginal_example_pendigits}
\end{figure}

\begin{table}[t]
	\centering
	\caption{Comparisons with deep clustering methods on the FashionMNIST dataset. The baseline results are taken from \citet{cai2022efficient} and ACC stands for clustering accuracy. 
	}
	\resizebox{\linewidth}{!}{
		\begin{tabular}{lcc}
			\hline\hline
			Method & ACC & NMI \\
			\hline
			K-Means \cite{bottou1994convergence} &  $0.474$ & $0.512$ \\
			DEC \cite{xie2016unsupervised} &  $0.590$ &  $0.601$\\
			IDEC \cite{guo2017improved} &  $0.592$  &  $0.604$\\
			VaDE \cite{jiang2017variational}  &   $0.578$  &  $0.630$\\
			JULE \cite{yang2016joint} &  $0.563$  &  $0.608$\\
			DAC \cite{chang2017deep}  &  $0.615$  &  $0.632$\\
			VaGAN-GMM \cite{yang2020clustering}  &  $\bf 0.638$ &   0.633\\
			EDESC \cite{cai2022efficient}  &   $\bf0.631$ &   $\bf0.670$\\
			\hline
			BigLearn-EM &   $0.624$ &   $\bf0.681$\\
			\hline\hline
		\end{tabular}
	}
	\label{table:deep_clustering}
\end{table}

The detailed results on the tested real-world clustering datasets are summarized in Table \ref{table:clustering}, where the BigLearn-EM delivers overall boosted performance over the compared techniques, especially on the NMI and ARI values.
When compared to the Joint-EM, the BigLearn-EM demonstrates significantly improved performance, even though both of them are based on E-M iterations; that further substantiates the effectiveness of the big learning principle in addressing the bad-local-optima challenge inherent in the vanilla Joint-EM algorithm; 
more importantly, the BigLearn-EM also delivers smaller standard deviations across the majority of tested datasets, demonstrating the potential of the big learning to bring better learning stability and consistency.
When compared to the WM-GMM and SW-GMM techniques that are developed based on complicated Wasserstein distances, the BigLearn-EM, which yields better performance, is clearly much easier to understand in theory and, simultaneously, easier to use in practice.

To probe the reasons underlying the boosted performance of the BigLearn-EM, we demonstrate the transformed marginal matchings associated with the top four principal components on the Pendigits dataset in Fig. \ref{fig:marginal_example_pendigits}, where, similar to the $25$-GMM simulation results in Fig. \ref{fig:JointEM_BiglearnEM_explicit}, the BigLearn-EM also delivers more accurate local matching to data modes on real-world datasets thanks to its big learning nature (see the arrows), manifested as its boosted performance on NMI/ARI values.


We also challenge the BigLearn-EM by comparing it with popular deep clustering methods. We follow the experimental setup in \citet{cai2022efficient} and conduct an experiment on the FashionMNIST dataset. The results are shown in Table \ref{table:deep_clustering}, where the BigLearn-EM delivers a comparable performance to SOTA deep clustering methods, without employing powerful deep neural networks, highlighting its effectiveness. 

\subsection{BigLearn-EM Is More Robust to the Scarcity of Its Training Data}


\begin{figure}[tbp]
    \centering
    \subfloat[NMI]{
        \includegraphics[width=0.45\columnwidth]{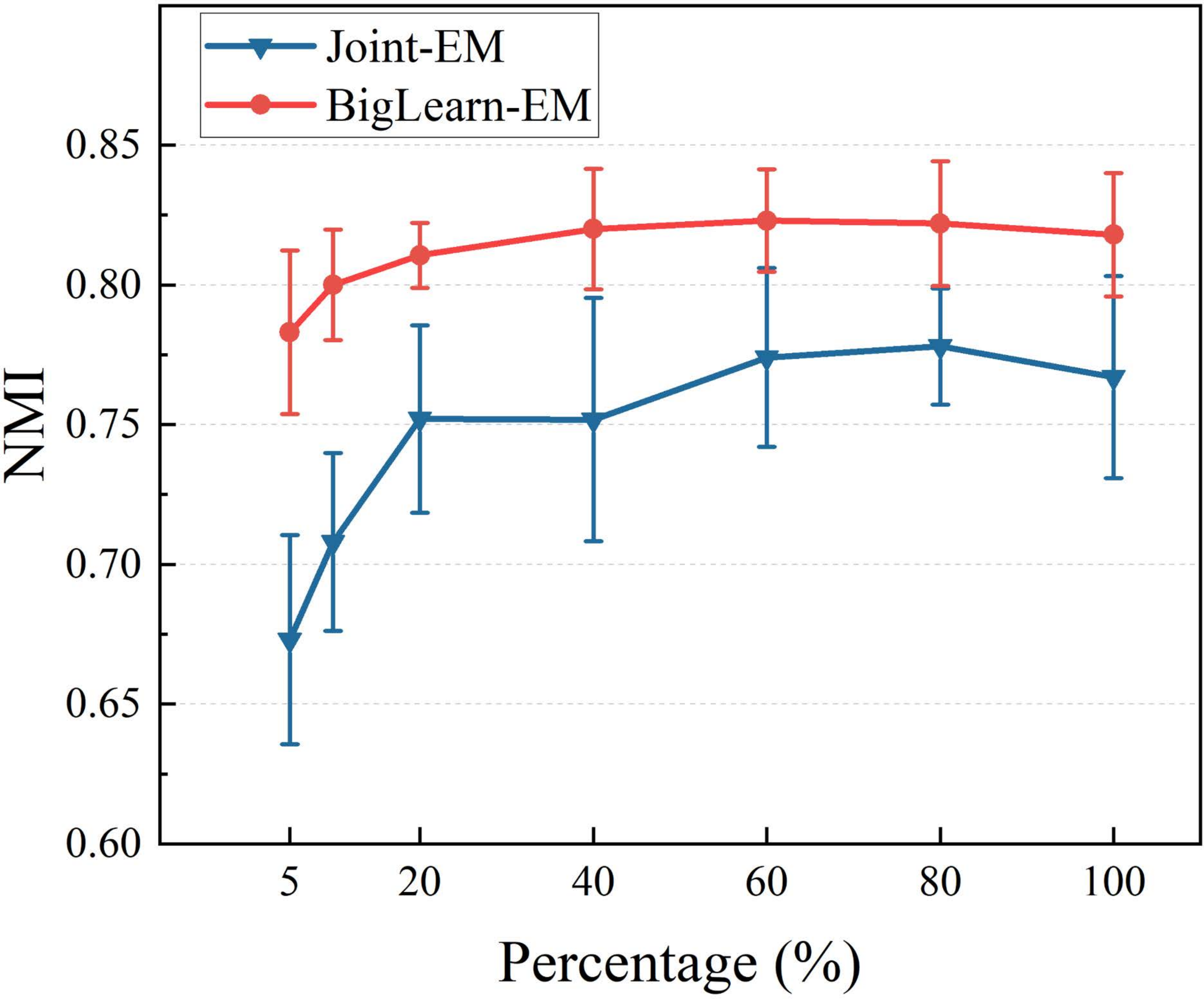}
        \label{fig:NMI_scarcity}}
    \,
    \subfloat[ARI]{
        \includegraphics[width=0.45\columnwidth]{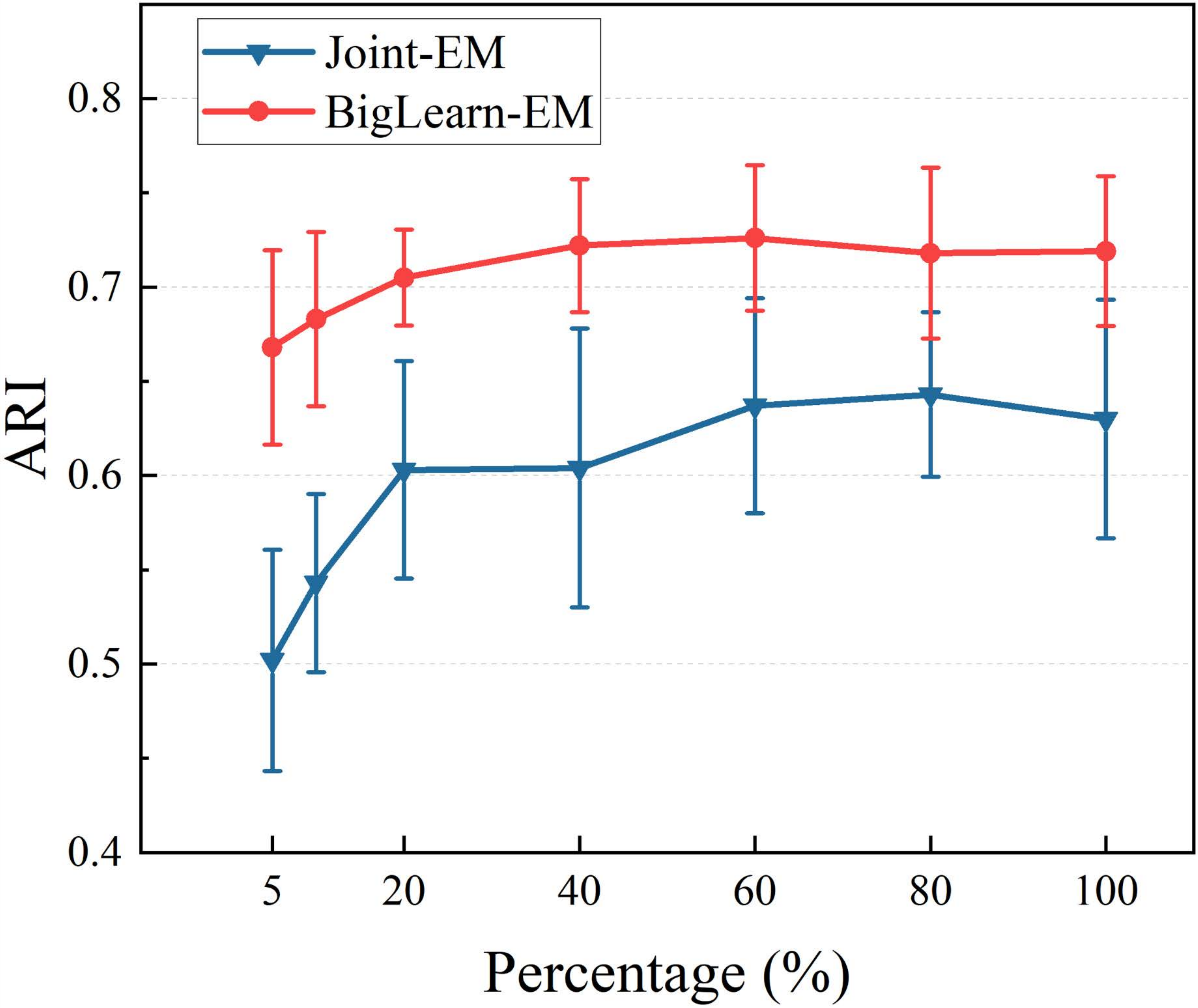}
        \label{fig:ARI_scarcity}}
    \caption{Demonstration of the BigLearn-EM's robustness to the scarcity of its training data.}
    \label{fig:exp_scarcity}
\end{figure}

Noticing that, in scenarios with limited data, like the Svmguide2 dataset with $391$ data samples in Table \ref{table:clustering}, the BigLearn-EM exhibits remarkably superior NMI/ARI performance than other clustering techniques.
We posit that the BigLearn-EM is more robust to the scarcity of its training data, because its big-learning operation is expected to significantly increase the utilization of the information within each sample.
We then design modified real-world clustering experiments to verify that hypothesis.
Specifically, based on the Pendigits dataset, we randomly select its $80\%$, $60\%$, $40\%$, $20\%$, $10\%$, and $5\%$ training data to form a series of modified clustering datasets with gradually increased scarcity, where the BigLearn-EM is compared to the Joint-EM to highlight the influence of the big learning principle.

Fig. \ref{fig:exp_scarcity} demonstrates the experimental results, which verify that the BigLearn-EM is more robust to the scarcity of its training data than the Joint-EM, even though both of them utilize similar EM-type parameter update formulas, highlighting the effectiveness of the big learning.





\section{Conclusions}

Leveraging the big learning principle that underlies groundbreaking foundation models, we upgrade the vanilla EM algorithm to its big-learning extension that is termed the BigLearn-EM.
The BigLearn-EM simultaneously performs joint, marginal, and randomly transformed marginal matchings between data and model distributions, empirically demonstrating great potential in addressing the long-lasting bad-local-optima challenge of the EM.
Comprehensive experiments on real-world clustering datasets demonstrate its boosted performance and its robustness to data scarcity.

Although the BigLearn-EM perform better than existing techniques in the tested scenarios, some issues remain unsolved. For example, 
($i$) whether the BigLearn-EM \emph{theoretically} addresses the bad-local-optima challenge of the EM is unanswered, 
($ii$) the consistency among joint, marginal, and transformed marginal matchings is not fully exploited, \eg to form a suitable stopping criteria for Algorithm \ref{alg:biglearn_EM},
and ($iii$) the exploration power of the BigLearn-EM may need further strengthening, as we find it may wander around a state like the one in Fig. \ref{fig:to_improve} for many iterations.


\section*{Acknowledgements}

We would like to thank the anonymous reviewers for their insightful comments. This work was supported in part by the Key Basic Research Project of the Foundation Strengthening Program (2023-JCJQ-ZD-123-06), the Natural Science Foundation of Guangdong Province, China, and the Pearl River Talent Recruitment Program (2019ZT08X751).

{\small
	\bibliography{ReferencesCong}
	\bibliographystyle{aaai24.bst}
}
\newpage


\appendix
\onecolumn

\begin{center}
    {\Large
        \textbf{
            Appendix of Big Learning Expectation Maximization
    }}
       

	\vspace{3mm}
	\textbf{Yulai Cong,
		\qquad\qquad 
		Sijia Li}
	
	Sun Yat-sen University\\
	yulaicong@gmail.com, lisijia57@163.com
    
\end{center}

\vskip 0.3in

\section{On Introducing the MAP Estimate of $\piv$}
\label{appsec:MAP_EM}

As the mixture weights $\piv$ is located in a simplex, \ie $\pi_i>0, \sum\nolimits_{i=1}^K \pi_i = 1$, a commonly used prior for $\piv$ is a Dirichlet distribution
\beq
p_{\alphav}(\piv) = \Dir(\piv; \alphav) = \frac{\Gamma(\sum\nolimits_{i=1}^K \alpha_i)}{\prod\nolimits_{i=1}^K \Gamma(\alpha_i)}  \prod\nolimits_{i=1}^K \pi_i^{\alpha_i - 1},
\eeq
where the concentration parameters $\alphav=(\alpha_1,\cdots,\alpha_K)$. Often, to encourage a full utilization of mixture components, one would prefer setting $\alpha_i > 1$. We set $\alpha_i = \alpha_j = \alpha > 1$ in this paper.

Taking the Joint-EM in \eqref{eq:EM} of the main manuscript as a demonstration example, we next elaborate on the the MAP Estimate of $\piv$, where the objective for $\piv$ is 
\beq\bali\label{eq:MAPloss_pi}
\gamma \log p_{\alphav}(\piv) + \Ebb_{q(\xv)}\log p_{\thetav}(\xv),
\eali\eeq
where $\gamma>0$ is a hyper-parameter that balances between the prior and the likelihood.
Note the first prior term is independent to $\{\muv_i, \Sigmamat_i\}_{i=1}^K$; therefore, the update rules for $q(z|\xv)$ and $\{\muv_i, \Sigmamat_i\}_{i=1}^K$ are the same as those of the Joint-EM in \eqref{eq:EM}.
One need only focus on the estimate of $\piv$.

The objective in \eqref{eq:MAPloss_pi} can be simplified \wrt $\piv$ as
\beq\bali
    \gamma {\log} p_{\alphav}(\piv) + \Ebb_{q(\xv)} {\log} p_{\thetav}(\xv) 
    & = C + \sum\nolimits_{i=1}^K \gamma(\alpha - 1) {\log} \pi_i + 
    \sum\nolimits_{z=1}^K \Ebb_{q(\xv)} [q(z|\xv)] {\log} \pi_z
    \\
    \st & \sum\nolimits_{z=1}^K \pi_z = 1
    \\
    & \pi_z \ge 0,
\eali\eeq
which is constrained optimization problem that can be readily solved by the method of Lagrange multipliers.
Accordingly, we have the MAP estimate of $\piv$ as
\beq\bali
    \pi^{*}_z = \frac{\Ebb_{q(\xv)} [q(z|\xv)] + \gamma(\alpha - 1)}{1 + \sum\nolimits_{z=1}^K \gamma(\alpha - 1)}
    = \frac{\Ebb_{q(\xv)} [q(z|\xv)] + \eta}{1 + K \eta}
\eali\eeq
where $\eta = \gamma(\alpha - 1) > 0$ and we conclude the derivation of \eqref{eq:post_pi} of the main manuscript.

\section{Settings of the Real-World Clustering Experiments}
\label{appsec:Exp_settings_cluster}




\textbf{Clustering Datasets} We adopt nine representative datasets\footnote{\url{https://www.csie.ntu.edu.tw/~cjlin/libsvmtools/datasets/multiclass.html}} that are extensively employed in the context of clustering, with their statistics summarized in Table \ref{clustering dataset}. 
We follow \citet{Chang2011LIBSVMsim} to normalize the data feature-wisely to the interval $[0, 1]$, using the min-max scaling.
For performance evaluation, we use the official testing data set if it's available; otherwise, we randomly select $20\%$ data samples to form a testing set.

\begin{table}[htb]
\centering
\caption{Statistics of the adopted real-world clustering datasets.}
\label{clustering dataset}
\begin{tabular}{cccc}
\hline
\hline
Dataset   & Dimension & Number & Class \\ \hline
Connect-4 & 126       & 67557  & 3     \\
Covtype   & 54        & 581012 & 7     \\
Glass     & 9         & 214    & 6     \\
Letter    & 16        & 20000  & 26    \\
Pendigits & 16        & 10992  & 10    \\
Satimage  & 36        & 6435   & 6     \\
Seismic   & 50        & 98528  & 3     \\
Svmguide2 & 20        & 391    & 3     \\
Vehicle   & 18        & 846    & 4     \\ \hline
\hline
\end{tabular}
\end{table}




\noindent\textbf{Performance Evaluation Metrics} 
We adopt three metrics for testing, that is,
\begin{itemize}[leftmargin=5mm]
    \item Normalized Mutual Information (NMI) \cite{strehl2002cluster}. The NMI score rescales mutual information scores using a generalized mean of the entropy of the true label set $\Omega$ and the cluster label set $C$. This process can be mathematically formulated as 
    \beq
	\operatorname{NMI}(\Omega, C)=\frac{I(\Omega ; C)}{(H(\Omega)+H(C)) / 2},
    \eeq
    where $I(\Omega ; C)=H(\Omega)+H(C)-H(\Omega, C)$ denotes the mutual information between $\Omega$ and $C$, and $H$ is the information entropy.
    
    \item Adjusted Rand Index (ARI) \cite{hubert1985comparing,steinley2004properties}. The ARI score represents an adjusted version of the Rand Index (RI) that accounts for chance. The RI itself serves as a measure of similarity, evaluating all possible pairs of samples and quantifying the instances where pairs are assigned to the same or distinct clusters in both predicted and true label assignments. The formalization of ARI is articulated as 
    \beq
        \mathrm{ARI}=\frac{(\mathrm{RI}-\text { Expected RI })}{(\max (\mathrm{RI})-\text { Expected } \mathrm{RI})}.
    \eeq
 
    \item  Test Joint Log-Likelihood (Joint-LL). The Joint-LL measures the extent to which the acquired model characterizes the testing dataset based on the joint KL divergence. It's calculated based on the testing data ${\xv}$ as 
    \beq
        \log p(\xv) = \log\sum\nolimits_{i=1}^K \pi_i \Nc(\xv|\muv_i, \Sigmamat_i).
    \eeq
	
\end{itemize}

\noindent\textbf{Parameter Settings}
The primary model parameters, \ie the number $K$ of mixture components, adopted for each datasets are listed in Table \ref{parameter configurations}.
For all the compared methods, we maintain the numerical stability by preventing the covariance matrices from being singular; specifically, we restrict the eigenvalues of each covariance matrix to be larger than hyperparameter $\epsilon >0$.
For the BigLearn-EM in Algorithm \ref{alg:biglearn_EM} of the main manuscript, to sample an index subset $\Tbb$ from the full index set $\Lbb$, we first sample a ratio $r_{\Tbb} \sim \text{Beta}(\beta_1, \beta_2)$ from the Beta distribution $\text{Beta}(\beta_1, \beta_2)$ with hyperparameters $(\beta_1, \beta_2)$; then, we uniformly sample $r_{\Tbb}$-ratio indices from $\Lbb$ to yield $\Tbb$; we run the BigLearn-EM for $10000$ iterations on all dataset and report the mean NMI, ARI, and Joint-LL values that are calculated based on the last $200$ iterations.


\begin{table}[H]
\centering
\caption{Parameter configurations.}
\label{parameter configurations}
\begin{tabular}{cccccccccc}
\hline\hline
Dataset & Connect-4 & Covtype & Glass & Letter & Pendigits & Satimage & Seismic & Svmguide2 & Vehicle \\ \hline
K       & $6$         & $10$      & $6$     & $26$     & $12$        & $8$        & $4$       & $3$         & $6$       \\
$\epsilon$     & $1\times 10^{-2}$      & $5\times 10^{-3}$    & $1\times 10^{-2}$  & $1\times 10^{-3}$   & $1\times 10^{-2}$      & $1\times 10^{-2}$     & $1\times 10^{-2}$    & $2\times 10^{-2}$      & $1\times 10^{-3}$    \\ \hline
$(\beta_1, \beta_2)$    & $(5,1)$     & $(5,1)$   & $(5,1)$ & $(5,1)$  & $(5,1)$     & $(5,1)$    & $(5,1)$   & $(5,1)$     & $(5,1)$   \\\hline\hline
\end{tabular}
\end{table}

\end{document}